\documentclass[final]{cvpr}

\usepackage{times}
\usepackage{epsfig}
\usepackage{graphicx}
\usepackage{amsmath}
\usepackage{amssymb}
\usepackage{multirow}
\usepackage{booktabs}
\usepackage{amsfonts,amssymb}
\usepackage{subcaption}
\usepackage[ruled,linesnumbered]{algorithm2e}
\usepackage[numbers,sort]{natbib}

\usepackage[pagebackref=true,breaklinks=true,colorlinks,bookmarks=false]{hyperref}

\usepackage{algorithmic}
\usepackage{soul}
\usepackage{xcolor}

\usepackage[capitalize]{cleveref}
\crefname{section}{Sec.}{Secs.}
\Crefname{section}{Section}{Sections}
\Crefname{table}{Table}{Tables}
\crefname{table}{Tab.}{Tabs.}

\def\cvprPaperID{7320} 

\def\confYear{2022}

\usepackage{url}            
\usepackage{booktabs}       
\usepackage{color}
\usepackage{multirow}

\usepackage[utf8]{inputenc} 
\usepackage[T1]{fontenc}    
\usepackage{hyperref}       
\usepackage{url}            
\usepackage{booktabs}       
\usepackage{amsfonts}       
\usepackage{subcaption}
\usepackage{graphicx}
\usepackage{color}
\usepackage{multirow}
\usepackage{amsmath}
\usepackage{amsfonts}

\begin{document}

\title{An Efficient Training Approach for Very Large Scale Face Recognition}

\author{
Kai Wang\textsuperscript{1,2}\thanks{Equal contribution. (\small{kai.wang@comp.nus.edu.sg, wangshuo514@sina.com})} 
\quad Shuo Wang \textsuperscript{2}\footnotemark[1]
\quad Panpan Zhang\textsuperscript{1}
\quad Zhipeng Zhou\textsuperscript{2}
\quad Zheng Zhu\textsuperscript{3}
\\
\quad Xiaobo Wang\textsuperscript{4}
\quad Xiaojiang Peng\textsuperscript{5}
\quad Baigui Sun\textsuperscript{2}
\quad Hao Li \textsuperscript{2}
\quad Yang You\textsuperscript{1}\thanks{Corresponding author (youy@comp.nus.edu.sg).}
\\
\textsuperscript{1}{National University of Singapore}
\quad \textsuperscript{2}{Alibaba Group}
\quad \textsuperscript{3}{Tsinghua University}
\\
\quad \textsuperscript{4}{Institute of Automation, Chinese Academy of Sciences}
\quad \textsuperscript{5}{Shenzhen Technology University}
\\
\small{Code: \url{https://github.com/tiandunx/FFC}}
}

\maketitle

\begin{abstract}
Face recognition has achieved significant progress in deep learning era due to the ultra-large-scale and well-labeled datasets.
However, training on the outsize  datasets is time-consuming and takes up a lot of hardware resource. Therefore, designing an efficient training approach is indispensable.
The heavy computational and memory costs mainly result from the million-level dimensionality of the fully connected (FC) layer.
To this end, we propose a novel training approach, termed Faster Face Classification (F$^2$C), to alleviate time and cost without sacrificing the performance. This method adopts Dynamic Class Pool (DCP) for storing and updating the identities' features dynamically, which could be regarded as a substitute for the FC layer. DCP is efficiently time-saving and cost-saving, as its smaller size with the independence from the whole face identities together.
We further validate the proposed F$^2$C method across several face benchmarks and private datasets, and  display comparable results, meanwhile the speed is faster than state-of-the-art FC-based methods in terms of recognition accuracy and hardware costs. Moreover, our method is further improved by a well-designed dual data loader including indentity-based and instance-based loaders, which makes it more efficient for updating DCP parameters. 
\end{abstract}
\section{Introduction}
\label{intro}
\begin{figure*}[htp]
\centering
\subfloat[Comparison of backbone and FC time cost (ms).]{\includegraphics[width = .45\linewidth]{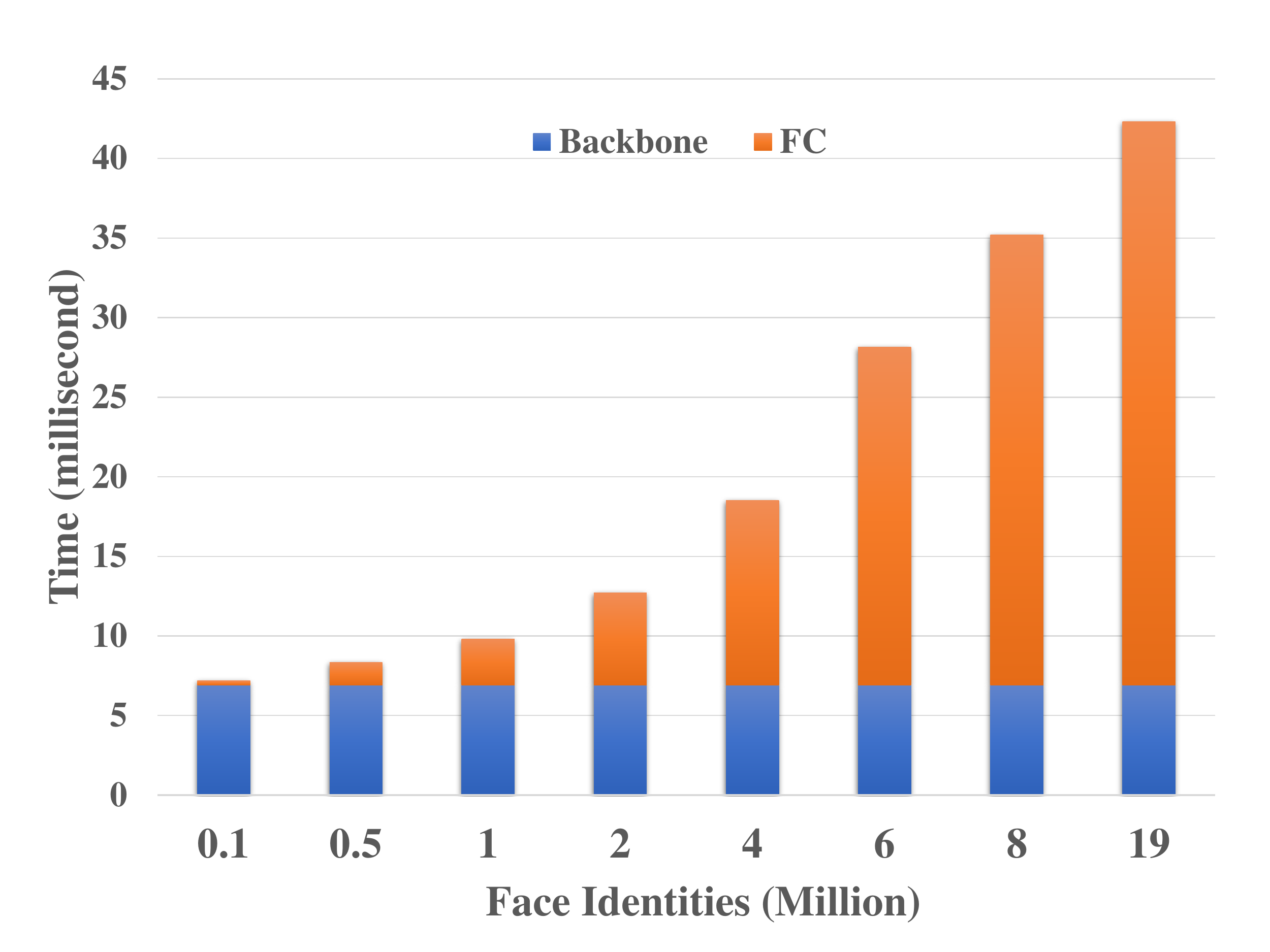}\label{fig:motivation_a}}\hfill
\subfloat[The memory occupancy of the FC layer at training phase (G).]{\includegraphics[width = .48\linewidth]{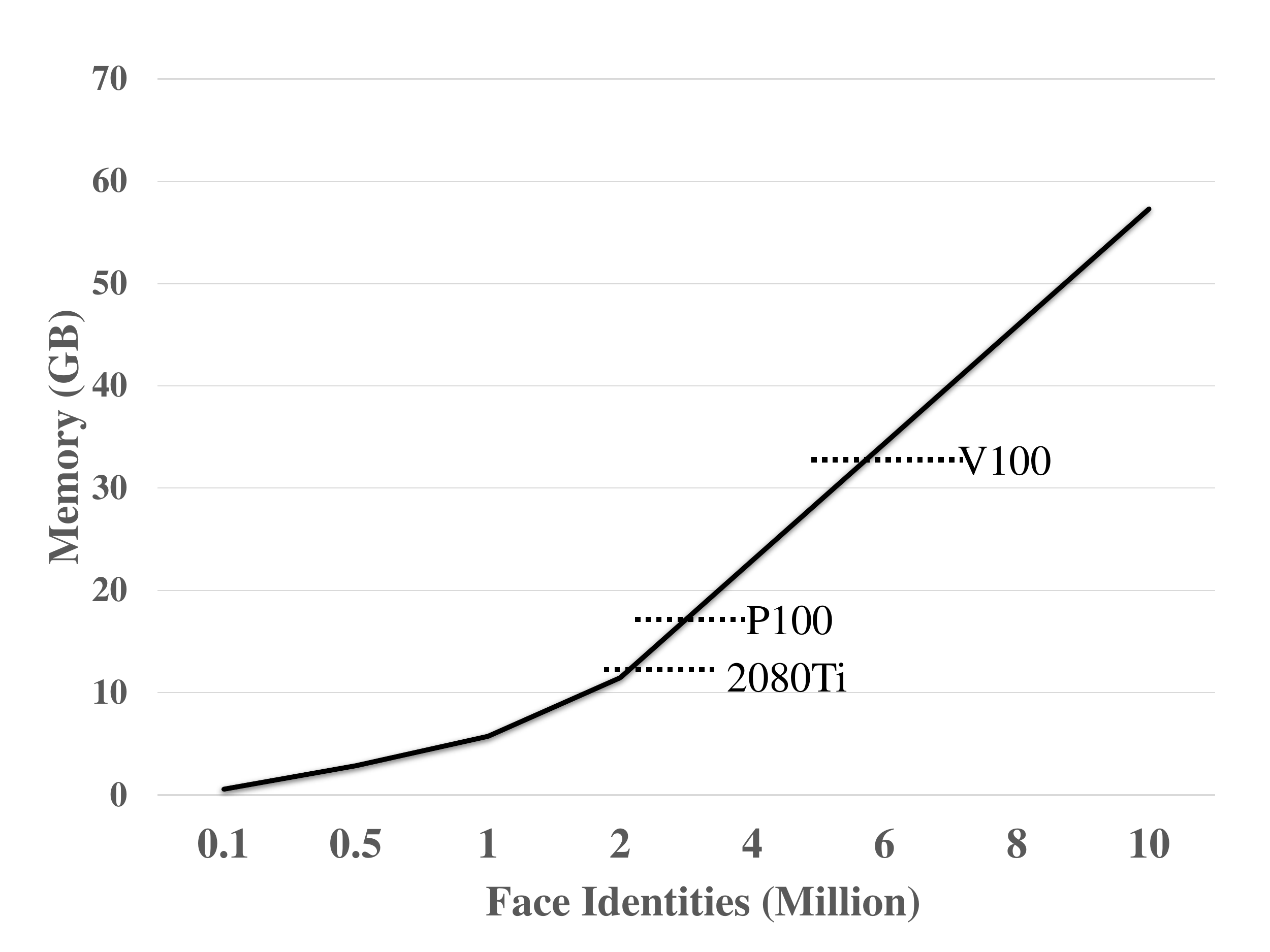}\label{fig:motivation_b}}
\caption{Visualization of training time and GPU memory occupancy.
Figure \ref{fig:motivation_a} shows the forward time comparison of backbone (ResNet50) and the FC layer.
Given an image, the time cost of FC increases sharply with the growing number of face identities but the time of backbone stays unchanged.
Figure \ref{fig:motivation_b} illustrates the GPU memory occupancy with the size of face identities.
Even the V100 32G GPU can only store the FC parameters with the output size of about 6 millions (The dimension of face recognition is usually 512).
Therefore, it is very necessary to design a method that reduces the training time and hardware cost of the FC layer.}     
\label{fig:motivation}
\end{figure*}

Deep Neural Networks (DNNs) has achieved many remarkable results in computer vision tasks \cite{peng2022crafting, wang2022cafe, wang2021mask, peng2020suppressing, wang2020region, wang2020suppressing}.
Face recognition can be regarded as one of the most popular research topics in computer vision.
Many large scale and well-labelled datasets have been released over the past decade \cite{yang2016wider, huang2008labeled, yi2014learning, guo2016ms, zhu2021webface260m}.
The training process of face recognition aims to learn identity-related embedding space, where the intra-class distances are reduced and inter-class distances are enlarged in the meanwhile. 
Previous works \cite{guo2016ms, wang2019co, wang2020loss} have proved that training on a large dataset can obtain a substantial improvement over a small dataset.
To this end, academia and industry collected ultra-large-scale datasets including 10 even 100 million face identities.
Google collected 200 million face images consisting of 8 million identities \cite{schroff2015facenet}.
Tsinghua introduced WebFace260M \cite{zhu2021webface260m} including 260 million faces, which is the largest public face dataset and achieves state-of-the-art performance.

In general, these ultra-large-scale datasets boost the face recognition performance by a large margin.
However, with the growth of face identities and limitations of hardware, there are mainly two problems in training phase.
The first problem results from the training time and hardware resource occupancy.
As shown in Fig \ref{fig:motivation}, the time cost and GPU memory occupancy of the FC layer are much greater than those of the backbone when the face identities reach 10 million.
To address these issues, many previous methods \cite{an2020partial, livirtual} focus on reducing the time and resource cost of the FC layer.
Previous methods can be summarized into two categories.
One \cite{an2020partial} tries to distribute the whole FC to different GPUs, introducing heavy communication costs. 
The other \cite{zhang2018accelerated} attempts to reduce the computing cost by selecting a certain ratio of neurons from the FC layer randomly, but it still needs to store the whole FC parameters. 
When the identities reach 10 or 100 million, storing the whole FC parameters is extremely expensive.
How to effectively reduce the computational and memory costs caused by the high-dimensional FC layer?
An intuitive idea is to decrease the size of FC or design an alternative paradigm, which is hardly explored before.
The second problem is related to the update efficiency and speed of FC parameters.
As pointed by \cite{du2020semi}, the optimal solution for the class center is actually the mean of all samples of this class.
Identities that have rare samples with very low frequency of sampling will have very little opportunity to updat class centers through their samples, which may hamper feature representation.

To tackle aforementioned issues, we propose an efficient training approach for ultra-large-scale face datasets, termed as Faster Face Classification (F$^2$C).
In F$^2$C, we first introduce twin backbones named Gallery Net (G-Net) and Probe Net (P-Net) to generate identity centers and extract face features, respectively.
G-Net has the same structure with P-Net and inherits the parameters from P-Net in a moving average manner.
Considering that the most time-consuming part of the ultra-large-scale training lies at the FC layer, we propose Dynamic Class Pool (DCP) to store the features from G-Net and calculate the logits with positive samples (whose identities appear in DCP) in each mini-batch.  DCP can be regarded as a substitute for the FC layer and its size is much smaller than FC, which is the reason why F$^2$C can largely reduce the time and resource cost compared to the FC layer.
For negative samples (whose identities do not appear in the DCP), we minimize the cosine similarities between negative samples and DCP.
To improve the update efficiency and speed of DCP parameters, we design a dual data loader including identity-based and instance-based loaders.
The dual data loader loads images from given dataset by instances and identities to generate batches for training.
Finally, we conduct sufficient experiments on several face benchmarks to prove F$^2$C can achieve comparable results and a higher training speed than normal FC-based method.
F$^2$C also obtains superior performance than previous methods in term of recognition accuracy and hardware cost.
Our contributions can be summarized as follows.

1) We propose an efficient training approach F$^2$C for ultra-large-scale face recognition training, which aims to reduce the training time and hardware costs while keeping comparable performance to state-of-the-art FC-based methods.


2) We design DCP to store and update the identities' features dynamically, which is an alternative to the FC layer.
The size of DCP is much smaller than FC and independent of the whole face identities, so the training time and hardware costs can be decreased substantially.

3) We design a dual data loader including identity-based and instance-based loaders to improve the update efficiency of DCP parameters.

\section{Related Work}
\label{relatedwork}
\textbf{Face Recognition.}
Face recognition has witnessed 
dramatical progress due to the large scale datasets, advanced architectures and loss functions.
Large scale datasets play the most crucial role in promoting the performance of face recognition \cite{deng2019arcface}.
These datasets can be divided into three intervals according to the number of face identities: 1-10K, 11-100K, >100K. VGGFace \cite{Parkhi15}, VGGFace2 \cite{8373813}, UMDFaces \cite{bansal2017umdfaces}, CelebFaces \cite{sun2015deeply}, and CASIA-WebFace \cite{yi2014learning} belong to the first interval.
The face identities of the IMDB-Face \cite{wang2018devil} and MS1MV2 \cite{deng2019arcface} are between 11K to 100K. Glint360k \cite{an2020partial} and Webface260M \cite{zhu2021webface260m} have about 0.36M and 4M identities. Many previous works \cite{an2020partial, zhang2018accelerated, zhu2021webface260m} illustrate that training on larger face identities datasets can achieve better performance than on smaller ones.
Therefore using WebFace260M as the training dataset obtains state-of-the-art performance on IJBC\cite{maze2018iarpa} and top 3 in NIST-FRVT challenge.
Based on these datasets, a variety of CNN architectures for improving the performances, such as VGGNet~\cite{simonyan2014very}, GoogleNet~\cite{szegedy2015going}, ResNet~\cite{he2016deep}, AttentionNet~\cite{wang2017residual} and MobileFaceNet~\cite{chen2018mobilefacenets}, have been proposed.
For the loss function, contrastive loss~\cite{sun2015deeply,yang2016large} and triplet loss~\cite{simonyan2014very} might be good candidates.
But they suffer from high computational cost and slow convergence.
To this end, researchers attempt to explore new metric learning loss functions to boost the face recognition performance.
Several margin-based softmax losses~\cite{liu2017sphereface,wang2018additive,wang2018support,wang2018ensemble,deng2019arcface} have been exploited and obtained the state-of-the-art results.
To sum up, current methods and large scale datasets have achieved excellent performance in face recognition, but the training time and hardware costs are still the bottleneck at training phase, especially for training on million scale or even more face identities datasets.

\textbf{Acceleration for Large-Scale FC Layer.}
As illustrated in Figure \ref{fig:motivation_a}, the time cost mainly focuses on FC layer rather convolutional layer when the face identities reach 10M.
Researchers try some attempts to accelerate the large scale FC training since 2001.
An intuitive idea is to design an approximate function to reduce the computational cost, the Hierarchical Softmax
(HSM)\cite{goodman2001classes} tries to reformulate the multi-class classifier into a hierarchy of binary classifiers.
Therefore, the training cost can be reduced by means of the given sample only has to traverse along a path from the root to the corresponding class.
However, all the class centers are stored in RAM and the retrieval time can not be ignored with the increase of face identities.
Zhang \textit{et.al.} \cite{zhang2018accelerated} proposed a method that can recognize a small number of \textbf{"active classes"} in each mini batch, which constructs the dynamic class hierarchies on the fly.
However, recognizing the \textbf{"active classes"} is also time-consuming when the face identities is too large.
Some companies, such as Google and Microsoft, try to divide all the categories into multi-GPUs averagely.
The communication cost of 
inter-servers can not be ignored.
To tackle this problem, Partial FC \cite{an2020partial} tries to train a large-scale dataset on a single GPU server using 10\% identities randomly at each iteration.
However it's still limited by the memory of the GPUs in a single machine.
As shown in Figure \ref{fig:motivation_b}, Partial FC can only work when the number of face identities is not ultra-large (<10M), otherwise the GPUs will still run out of memory.
There are several pairwise based methods \cite{kang2018pairwise} that utilize the face pairs to train large scale datasets, while the time complexity is $O(N^k)$, where $k$ represents the size of the pair.
The latest related work VFC \cite{livirtual} builds some virtual FC parameters to reduce the computation cost but its performance is much lower compared to normal FC.
Different from previous works, our F$^2$C can reduce the FC training cost largely and achieve comparable performance compared to normal FC-based methods.


\section{Faster Face Classification}
\label{method}

In this section, we first give an overview of F$^{2}$C for a brief understanding of our method.
Then we present our motivation and key modules for ultra-large-scale datasets training.
After that, we show the theoretical/empirical analysis over these modules.
Finally we demonstrate the training details for better reproduction.

\begin{figure*}[htp]
\centering
\includegraphics[width=0.8\textwidth]{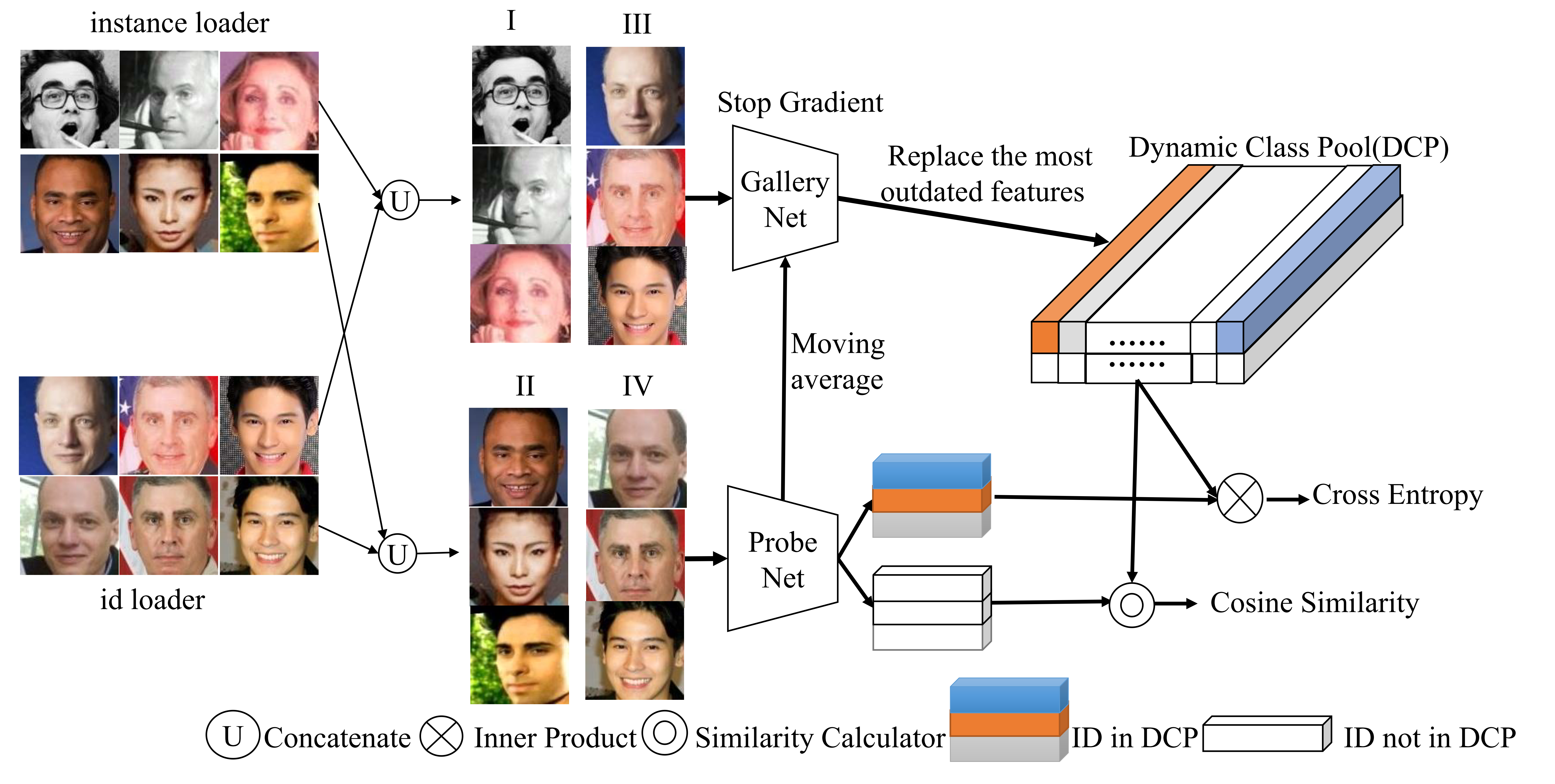}
\caption{The pipeline of F$^2$C. We use instance and id data loader to generate mixed batches (I $\cup$ III, II $\cup$ IV), which are later fed into G-Net and P-Net respectively. The features from G-Net will update DCP in the manner of LRU, and features from P-Net will be used to compute loss together with DCP.}

\label{pipeline}
\end{figure*}

\subsection{Overview of F$^{2}$C}
The problem we tackle is to accelerate the training speed and reduce the hardware costs of ultra-large-scale face datasets (face identities > 10M) without obvious degradation of performance. 
To this end, we propose F$^{2}$C framework for ultra-large-scale face datasets training.
As shown in Figure \ref{pipeline}, given ultra-large-scale face datasets, we utilize instance-based loader to generate an instance batch as data loader usually does. Meanwhile, identity-based loader selects two images randomly from the same identity to form the paired identities batch.
Subsequently, we mix up the images from instance and pair identity batches as shown in Figure \ref{pipeline} and feed them into G-Net and P-Net.
Inspired by MoCo \cite{he2020momentum}, G-Net has the same structure as P-Net and inherits parameters from P-Net in a moving average manner.
G-Net and P-Net are used to generate identities' centers and extract face features for face recognition, respectively.
Then, we introduce DCP as a substitute for the FC layer.
DCP is randomly initialized and 
updated by the features from G-Net at each iteration.
The update strategy of DCP follows the rule: using the current features to replace the most outdated part of features in DCP. For positive samples, we use the common cross entropy loss. For negative samples, we minimize the cosine similarities between negative samples and DCP. The whole F$^{2}$C is optimized by cross entropy loss and cosine similarities simultaneously.

\subsection{Motivation}
Before digging into F$^{2}$C, we provide some motivations by rethinking the loss function cooperated with FC layer.
For convenience, we consider the Softmax as follows:
\begin{equation}\label{softmax}
   L= -\frac{1}{N}\sum_{i=1}^{N}\log\frac{e^{W_{y_i}^Tx_i}}{\sum_{j = 1}^{n_{\text{ID}}}e^{W_{j}^Tx_i}}
\end{equation}
where $N$ is batchsize and $n_{\text{ID}}$ stands for the number of whole face identities.
For each iteration of the training process, the update of the classifier $\{W_{j}\}_{j=1}^{n_{\text{ID}}}$ is performed as the following equations:
\begin{equation}\label{update_W}
\frac{\partial L}{\partial W_{k}}=
    \begin{aligned}
    &-\frac{1}{N}\sum_{i=1}^{N}(\delta_{ky_{i}}-\frac{e^{W^{T}_{k} x_{i}}}{\sum_{j=1}^{n_{\text{ID}}}e^{W^{T}_{j} x_{i}}})x_{i}
    \end{aligned}
\end{equation}
Obviously, all the classifiers $\{W_{j}\}_{j=1}^{n_{\text{ID}}}$ will be updated in each iteration, which means each classifier has the same chance to be optimized.
The goal of face recognition is to distinct persons from different identities with the mechanism where the features from the same identity are pulled together and features belonging to different identities are pushed away.
As the main problem of training with ultra-large-scale dataset is the explosive size of FC layer,
We can consider the whole FC as a set of classifiers.
In order to reduce the computation cost, it is intuitive for us to optimize fixed ratio of the classifiers in each iteration during the training process.
Specifically, we utilize a vector as follows to represent whether a given classifier is in optimization queue.
\begin{equation}\label{V_setting}
    V = \{\nu_{1},...,\nu_{n_{\text{ID}}}\}, \forall i, \nu_{i}\in\{0,1\} \text{ and } \#\{\nu_{i}|\nu_{i}\neq 0\}=C
\end{equation}
where $C$ is a constant stands for the length of the optimization queue, $\nu_{i}=0/1$ denotes the classifier $W_{i}$ is (not in)/(in) optimization queue.
We draw the corresponding objective for this setting.
\begin{equation}\label{part_loss}
\hat{L}=-\frac{1}{N}\sum_{i=1}^{N}\log\frac{e^{W^{T}_{y_{i}} x_{i}}}{\sum_{j=1}^{n_{\text{ID}}}\nu_{j}e^{W^{T}_{j} x_{i}}}
\end{equation}
The classifiers update on basis of the following equations: 
\begin{equation}\label{update_part_W}
\frac{\partial \hat{L}}{\partial W_{k}}=
    \begin{aligned}
    &-\frac{1}{N}\sum\limits_{i=1}^{N}(\delta_{ky_{i}}-\frac{\nu_{k}e^{W^{T}_{k} x_{i}}}{\sum_{j=1}^{n_{\text{ID}}}\nu_{j}e^{W^{T}_{j} x_{i}}})x_{i}
    \end{aligned}
\end{equation}
Formally, equation \ref{update_part_W} is similar to equation \ref{update_W}, the selection mechanism for vector $V$ will influence the update process of the classifier directly.
We should design feasible selection mechanism for better optimization of classifiers under the constraint that only partial classifiers will be updated at each iteration.
However, this straightforward method still suffer the heavy pressure from storing the whole set of classifiers.
As a matter of fact, in our novel framework, we only offer limited space to store a fixed ratio of classifier/features dynamically.
\subsection{Identity-Based and Instance-Based Loaders}\label{dual_loader}
In this subsection, we introduce the details of our dual data loader.
For convenience, we denote the batchsize as \textbf{M}.
Practically, we utilize the instance-based loader to sample \textbf{M} images from a given face dataset randomly to get the instance batch.
In the meanwhile, the identity-based loader is applied to provide identity batch by selecting \text{M} identities randomly without replacement from the whole identities and sample two images for each identity.
We divide the instance batch into two parts, with \textbf{M/2} images for each part.
For paired identity batch, we split it by face identity to form two parts with same set of face identities.
We mix up the four parts to get $ I \cup III; II \cup IV$ (as illustrated in Figure \ref{pipeline}), where
$\cup$ represents the union operation for sets. 

\textbf{Why Dual data loaders?}
As aforementioned, we design dual data loader to improve the update efficiency of DCP parameters.
To better understand our design, we analyze the different influences between identity-based and instance-based loaders as follows.
Let $M$ denote batch size, $n_{ID}$ be the total number of identities of the given the dataset, $k_{min}$ ($k_{max}$) as the minimum (maximum) number of images for one person in the dataset, $\bar{k}$ be the average number of images per identity.
Here the shape of DCP mentioned in the main paper is $C \times K \times D$.
$C$ is the magnitude of DCP, $K$ is the capacity for each placeholder in DCP, $D$ represents feature dimension.
The total images of given dataset can be denoted as $\bar{k}n_{ID}$.
We evaluate the update speed by estimating the minimum of epochs for  given face identity to update $\frac{\bar{k}n_{\text{ID}}}{M}$.
\begin{itemize}
\item If we only use instance-based loader, the update speed of identities' centers $\in[\frac{\bar{k}n_{ID}}{Mk_{max}}, \frac{\bar{k}n_{ID}}{Mk_{min}}]$. So only using instance-based loader may lead to following problems.
1. If the number of identities is severely imbalanced, the update speed of the identities' centers that have rare number of images is too slow.
2. If we sample $M$ images that belong to $M$ different identities, the DCP may have no positive samples for this iteration.
In this case, cross entropy, which is crucial for classification, cannot be calculated.

\item If we only use identity-based loader, we can obtain the average fastest update speed ($\frac{n_{ID}}{M}$) of each identity.
However, identity-based loader re-sample identities that have rare number of images too many times, so it needs to use about $\frac{k_{max}}{k_{min}}$ times more iterations than instance-based loader to sample all images from the dataset.
Further, the sample probabilities for each instance of identities with rich intra-class images are too low, the identity-based loader can not sample plenty of intra-class images during the training phase.

\item Using the dual data loaer can inherit the benefits from instance-based and identity-based loaders.
First, dual data loader provides appropriate ratios between positive and negative images, which is very important for DCP.
Second, dual data loader keeps high update efficiency (speed) of identities' centers and various intra-class images.
\end{itemize}

\textbf{Feature Extraction} We take $I \cup III$ and $II \cup IV$ as input to Probe and Gallery Nets respectively to extract the face features and generate the identities' centers.
The process can be formulated as follows:
\begin{equation}\label{feature_P_G}
    \begin{aligned}
    &P_{\theta}(I\cup III)=F^{\text{DCP}}_{p}\oplus F^{\neg\text{DCP}}_{p}\\
    &G_{\phi}(II\cup IV)=F_{g}
    \end{aligned}
\end{equation}
where the probe and gallery net are abbreviated as $P_{\theta}$ and $G_{\phi}$ with parameters denoted as $\theta,\phi$ respectively. 
The symbol $\neg$ is set to split features whose identities belong to DCP (subsection \ref{DCP_sec}) from those not belong to DCP.
$F_{g}$ represent the features extracted by the Gallery Net.
For each batch, we denote number of identities in DCP as $I$ and number of identities not in DCP as $M-I$.

\subsection{Dynamic Class Pool}\label{DCP_sec}
In this subsection, we introduce the details of the Dynamic Class Pool (DCP). 
Inspired by sliding window \cite{lampert2008beyond} in object detection task, we can utilize a dynamic identity group that slides the whole face identities by iterations.

We called this sliding identity group as DCP, which can be regarded as a substitute for the FC layer.
Firstly, we define a tensor $T$ with size of $C \times K \times D$ which is initialized with Gaussian distribution,
where $C$ is the capacity or the number of face identities the DCP can hold,
$K$ represents the number of features that belong to the same identity (we set the default as $K=2$).
We store $F_g$ in DCP and update the most outdated features of DCP using the $F_g$ in each iteration.
The updating rule is similar to least recently used (LRU)\footnote{https://www.interviewcake.com/concept/java/lru-cache} policy which can be formulated as,
\begin{equation}
\label{DCP}
    \begin{aligned}
    &T[1:C-M,:,:]=T[M+1:C,:,:]\in\mathbb{R}^{(C-M)\times K\times D}\\
    &T[C-M+1:C,0,:]=F_{g}\in\mathbb{R}^{M\times K\times D}
    \end{aligned}
\end{equation}
For the current batch, with the update of the DCP, we obtain pseudo feature center for each identity in DCP, including the identities contained in $II\cup IV$.
As claimed in equation \ref{feature_P_G}, features from P-Net can be divided into two types compared to DCP.
One is $F_{p}^{\text{DCP}}$, the other is $F_{p}^{\neg\text{DCP}}$.
For $F_{p}^{\text{DCP}}$, we can calculate its logits by the following equation,
\begin{equation}
    P = \frac{1}{K}\sum_{i=1}^{K}\left< F_{p}^{\text{DCP}},T[:,i,:] \right> \in \mathbb{R}^{I \times C}
\end{equation}
where $\left< \cdot,\cdot \right>$ denotes the inner product operation, $P$ represents the logits of $F_{p}^{\text{DCP}}$.
Therefore, we can formulate the Cross-Entropy loss as follows:
\begin{equation}\label{ce_loss}
L_{\text{ce}} = -\frac{1}{I}\sum_{i=1}^{I}\log\frac{e^{W_{y_i}^{\mathrm{T}}P_i}}{\sum_{j=1}^{C}e^{W_{j}^{\mathrm{T}}P_i}},
\end{equation}
where $W_j$ is the j-th classifier, $y_i$ is the identity of $P_i$.
For features $F_{p}^{\neg\text{DCP}}$ whose IDs are not in DCP, we add a constraint to minimize the cosine similarity between $F_{p}^{\neg\text{DCP}}$ and $T$, which can be formulated as,
\begin{equation}
    L_{\text{cos}} = \frac{1}{M-I}\sum_{i=1}^{M-I}\varphi(F_{p}^{\neg\text{DCP}}, \bar T),
\end{equation}
where $\varphi$ is the operation of calculating the cosine similarity, $\bar T$ represents the average operation along the axis of $K$ in DCP. 
The total loss is $L_{\text{total}}$ = $L_{\text{ce}}$ + $L_{\text{cos}}$.

\subsection{Empirical Analysis}
\textbf{DCP}
As shown in equation \ref{part_loss} and \ref{ce_loss}, the cross entropy loss we utilize for DCP is similar to the loss for FC formally.
With special setting of vector $V$ in equation \ref{V_setting}, we can represent $L_{\text{ce}}$ in the form of equation \ref{part_loss}. 
For further verification of the effect of this mechanism on the training with DCP, we provide some empirical analysis.



\begin{algorithm}[htp]
\caption{Update Mechanism of DCP}
\label{FIFO_kai}
\KwIn{

\noindent DCP: $T\in \mathbb{R}^{C\times K\times D}$ initialized with Gaussian distribution.

\noindent Index for the identity batch: $t$.

\noindent Batch Size: $M$.
}
\For{$1\le t \le \frac{n_{ID}}{M}$}{
utilize the G-Net to extract features from $t$-th batch as the pseudo feature centers denoted as $F_{g}$;\\
\textbf{if} $1\le t \le \frac{C}{M}$:\\
\quad store $F_{g}$ sequentially in those unoccupied position in DCP.\\
\textbf{else}:\\
update DCP as shown in Equation \ref{DCP}}

\end{algorithm}

As mentioned in subsection \ref{dual_loader} and equation \ref{DCP}, the identities in DCP are updated in an LRU mechanism as shown in Algorithm \ref{FIFO_kai}.
As identity-based loader goes through the dataset in terms of identities,
partial components($\frac{M}{2}$) of vector $V$ can be determined by shuffling the whole face identities and taking the corresponding $t$-th part of it, where $1\le t\le \frac{n_{ID}}{M}$.
When we use identity-based loader,
then by the setting of $V$ and property of LRU rules, each classifier/pseudo feature center can be updated at least $[\frac{C}{M}]$ times.
This means that every classifier can have the similar chance to be optimized in our settings.
DCP may have the following benefits: 1) The size of DCP is independent from magnitude of face identities, which can be far smaller than FC. Therefore the computational cost is greatly reduced; 2) The hardware especially storage occupancy of DCP is also smaller than FC and the communication cost can be reduced dramatically.
These benefits are the reasons why we call our method as Faster Face Classification.

\subsection{Experimental Details}
We train our F$^2$C on a single server with 8 Tesla V100 32G GPUs.
We utilize ResNet100, ResNet50 and MobileFaceNet as our backbones to evaluate the efficiency of F$^2$C.
The learning rate is initialized as 0.1 with SGD optimizer and divided by 10 at 10, 14, 17 epochs.
The training is terminated at 20 epochs.
The length (number of ID) of DCP is defaulted as 10\% of total face identities.
The batch size is 512 \textit{i.e.}, 256 images from identity-based loader and 256 images from instance-based loader. 


\section{Experiments}
\label{experiments}

\begin{table*}[htp]
\center
\caption{Evaluation results (\%) on 9 face recognition benchmarks. All models are trained from scratch on MS1MV2, Glint360k and Webface42M. The TPR@FAR=1e-4 metric is used for IJBC. MegaFace is TPR@FAR=1e-6}
\scalebox{0.8}{
\resizebox{\linewidth}{!}{
\begin{tabular}{@{}ccccccccccccccc@{}}
\toprule
Method & LFW &SLFW &CFP &CALFW &CPLFW &AGEDB &YTF  &IJBC & MegaFace &Avg. \\ 

	&&&&\multicolumn{3}{l}{\textbf{Training on MS1MV2}}\\
	\midrule
	FC-Mobile   &99.04 &98.80 &96.94 &94.37 &88.37 &96.73 &97.04 &92.29  & 90.69 &94.92\\
	F$^2$C-Mobile  &98.93&98.57 &97.16 &94.53 &87.80 &96.47 &97.24 &91.06  & 89.30 &\textbf{94.56}\\

FC-R50  &99.78 &99.55 &98.80 &95.76 &92.01 &98.13 &98.03 &95.74  &97.82  &97.29\\
F$^2$C-R50   &99.50 &99.45 &98.46 &95.58 &90.58 &97.83 &98.16 &94.91  &96.74  &\textbf{96.80}\\

	&&&&\multicolumn{3}{c}{\textbf{Training on Glint360k}}\\
\midrule
FC-R50   &99.83 &99.71 &99.07 &95.71 &93.48 &98.25 &97.92 &96.48  & 98.64 &97.67\\
F$^2$C-R50  &99.71 &99.53 &98.30 &95.23 &91.60 &97.88 &97.76 &94.75  &96.73 &\textbf{96.83}\\

	&&&&\multicolumn{3}{c}{\textbf{Training on Webface42M}}\\
	\midrule
	FC-R100 & 99.83 & 99.81 & 99.38 & 96.11 & 94.90 &98.58 & 98.51 & 97.68 & 98.57 & 98.15 \\
	F$^2$C-R100 & 99.83 & 98.80 & 99.33 & 95.92 & 94.85 & 98.33 & 98.23 & 97.31 & 98.53 & \textbf{97.90} \\


\bottomrule
\end{tabular}}}
\label{tab:1}
\end{table*}

In this section, we first review several benchmark datasets in face recognition area briefly.
Then, we conduct ablation studies to evaluate the effectiveness of each module and the settings of hyper-parameters in F$^2$C.
Finally, we compare F$^2$C to related state-of-the-art methods.

\subsection{Datasets.} 

We utilize MobileFaceNet, ResNet50 and ResNet100 to train F$^2$C on MS1MV2, Glint360k and Webface42M( Webface42M is the cleaned version of the original Webface260M and it has 2M ID and about 42M images), respectively. We mainly show the performances of F$^2$C in following 9 academic datasets: LFW~\cite{huang2008labeled}, SLFW~\cite{huang2008labeled}, CFP~\cite{7477558}, CALFW~\cite{zheng2017cross}, CPLFW~\cite{zheng2018cross}, AGEDB~\cite{moschoglou2017agedb}, YTF~\cite{wolf2011face}, IJBC~\cite{maze2018iarpa}, and MegaFace~\cite{kemelmacher2016megaface}. LFW is collected from the Internet which contains 13,233 images with 5,749 IDs.
SLFW is similar to the LFW but the scale of SLFW is smaller than LFW.
CFP collects celebrities' images including frontal and profile views.
CALFW is a cross-age version of LFW.
CPLFW is similar to CALFW, but CPLFW contains more pose variant images.
AGEDB contains images annotated with accurate to the year, noise-free labels.
YTF includes 3425 videos from YouTube with 1595 IDs.
IJBC is updated from IJBB and includes 21294 images of 3531 objects.
MegaFace aims at evaluating the face recognition performance at the million scale of distractors, which includes a large gallery set and a probe set.
In this work, we use the Facescrub as the probe set of MegaFace as gallery.

\subsection{Performance Comparisons between FC and F$^2$C}
We choose 3 different backbones and evaluate the performance on 9 academic benchmarks between FC and F$^2$C using MS1MV2, Glint360k and Webface42M as training datasets.
As shown in Table \ref{tab:1}, F$^2$C can achieve comparable performance compared to FC.
We also provide the average performance among these datasets and demonstrate it in the last column  where F$^2$C is only lower than FC within 1\%.
Note that, the size of DCP is only 10\% of the total face identities.

\subsection{Ablation Studies}
We conduct ablation studies of hyper parameters and settings of F$^2$C.
Here we demonstrate the experiments on MS1MV2 using MobileFaceNet and ResNet50. 

\textbf{Single Loader or Dual Loaders?} As mentioned in methodology section, dual loaders can improve the update efficiency of DCP.
To evaluate the influence of loaders in F$^2$C, we use different combinations of identity-based and instance-based loaders and show the results in Table \ref{tab:2}.
The Small Datasets represent LFW, SLFW, CFP, CALFW, CPLFW, AGEDB and YTF in this subsection. We show the average accuracy on Small Datasets. Unless specified, TPR@FAR=1e-4 metric is used for IJBC and Megafce is FPR@FAR=1e-6 by default.
Training with instance-based loader or identity-based loader can obtain comparable results on small datasets.
Instance-based loader outperforms identity-based loader on IJBC and MegaFace by a large margin.
It could be explained that only using identity loader can not ensure all the images are sampled.
Using dual data loaders can improve the performance compared with each single loader obviously, which is consistent to our analysis. Note that, to make fair comparison, the results are obtained with the same number of samples fed to the model, not with the same number of epoch.

\textbf{Single Net or Dual Nets?} MoCo treates the two augmented images of the same image as positive samples and achieved impressive performance in unsupervised learning. Therefore, pictures with the same ID can naturally be regarded as positive samples, thus it is intuitive to use twin backbones in the same way as MoCo to generate the identities' centers and extract the face features respectively. However we intend to reduce the training cost further, so we compare the performance of single net to dual nets in Table \ref{tab:3}. \textcolor{black}{The dual nets performs better than single net on all the datasets, which illustrates only using single net may fall into the trivial solution as explained in Semi-Siamese Training\cite{du2020semi}.}

\begin{table}[t]
\centering
\caption{Evaluation of single or dual data loaders.ID.L, Ins.L and Dua.L represent id loader, instance loader and dual loaders respectively.}
\label{tab:2}
\setlength{\tabcolsep}{1mm}{
\begin{tabular}{@{}ccccc@{}}
\toprule
Backbone & Method  &Small Datasets  &IJBC & MegaFace \\ 
	
	\midrule
	 & ID.L   &94.20  &82.30 &79.19\\
	Mobile & Ins.L   &94.24  &89.30 &86.40\\
	 & Dua.L  &\textbf{95.29}  &\textbf{91.06} &\textbf{89.30}\\
	 \hline
	 & ID.L & 96.70 & 91.75 & 93.65 \\ ResNet50 & Ins.L & 96.08 & 92.06 & 92.74 \\ & Dua.L & \textbf{97.07} & \textbf{94.91} & \textbf{96.74}\\

\bottomrule
\end{tabular}} 
\end{table}

\begin{table}[t]
\centering
\caption{Evaluation of single net or dual nets.}
\label{tab:3}
\vspace{-0.5em}
\setlength{\tabcolsep}{1mm}{
\begin{tabular}{@{}ccccccc@{}}
\toprule
Backbone &Method  &Small Datasets   &IJBC & MegaFace \\ 
	
	\midrule
	\multirow{2}[2]*{Mobile} & Single   &93.90  &88.07 &82.69\\
	 & Dual &\textbf{95.29}  &\textbf{91.06} &\textbf{89.30}\\
	 \hline
	 \multirow{2}[2]*{ResNet50}& Single& 95.55 & 92.26 & 92.98 \\
	 &Dual & \textbf{97.07} & \textbf{94.91} & \textbf{96.74} \\

\bottomrule
\end{tabular}}
\end{table}


\textbf{Exploring the Influence of K in DCP.} $K$ represents the number of the features that belong to the same identity. We evaluate the $K=1$ and $K=2$ in Table   \ref{tab:4}.
As the features in DCP represent the category centers, an intuitive sense is that a larger $K$ can provide more reliable center estimation.
The experiments results also support our intuition.
However we must make a trade-off between performance and storage.
A larger $K$ means better performance at the cost of GPU memory and communication among severs.
Therefore, we set $K=2$ in DCP by default.

\begin{table}[t]
\centering
\caption{Evaluation of the number of K.}
\label{tab:4}
\vspace{-0.5em}
\setlength{\tabcolsep}{1mm}{
\begin{tabular}{@{}ccccc@{}}
\toprule
Backbone & K  &Small Datasets  &IJBC & MegaFace \\ 
	
	\midrule
	\multirow{2}[2]*{Mobile} & 1   &95.19  &90.75 &88.31\\
	&2 &\textbf{95.29}  &\textbf{91.06} &\textbf{89.30}\\
	\hline
	\multirow{2}[2]*{ResNet50} & 1 & 96.58 & 94.38 & 96.49\\
	&2 & \textbf{97.07} & \textbf{94.91} &\textbf{96.74} \\

\bottomrule
\end{tabular}} 
\end{table}

	


	


	


\begin{table}[t]
\centering
\caption{Evaluation the ratios within dual data loader. ResNet50 is used here.}
\label{tab:5}
\vspace{-0.5em}
\setlength{\tabcolsep}{1mm}{
\begin{tabular}{@{}cccccccc@{}}
\toprule
Ins.L & ID.L & Small Datasets  &IJBC & MegaFace \\ 
	
	\midrule
	0 & 1 & 96.77 & 91.75 &93.65\\
	1 & 0 & 96.23 & 92.06 &92.74\\
	1 & 1 & \textbf{97.08} & \textbf{94.91} & \textbf{96.74}\\
	2 & 1 & 96.29 & 94.21 & 96.43 \\
	1 & 2 & 95.40 & 90.80 & 90.56\\

\bottomrule
\end{tabular} }
\end{table}

\begin{table*}[tp]
    \centering
    \caption{Comparisons to state-of-the-art methods.
    To make fair comparison, Partial-FC, VFC, DCQ and \textbf{F$^2$C} only use 1\% of identities of MS1M for training.
    Megaface refers to rank-1 identification. IJBC is TPR@FAR=1e-4.
    The lower-boundary results are excerpted from VFC paper.
    The upper-boundary results are reproduced by us.}
    \begin{tabular}{@{}ccccccccccccccc@{}}
\toprule
Method  &CALFW &CPLFW &SLFW &YTF   &CFP &IJBC & MegaFace \\ 
	
	\midrule
	lower-boundary &87.43  &75.45 &93.52 &93.78 &91.66 &65.19 &79.28\\
	upper-boundary &95.75  &90.85 &99.55 &97.76  &98.39 &95.48 &97.56\\
	N-pair\cite{sohn2016improved} &87.32  &72.80 &92.28  &92.62  &- &61.75 &82.56\\
	Multi-similarity\cite{wang2019multi} &85.40  &73.60 &91.03 &92.76  &- &57.82 &76.88\\
	TCP\cite{liu2018transductive} &88.05  &76.00 &93.23 &93.92  &93.27 &43.58 &88.18\\
	Partial-FC\cite{an2020partial}   &95.40  &90.33 &99.28 &97.76  &98.13 &94.40 &94.13\\
	VFC\cite{livirtual}   &91.93  &79.00 &96.23 &95.08  &95.77 & 70.12 &93.18\\
	DCQ\cite{li2021dynamic}& 95.38 & 88.92 & 99.23 & 97.71 & 98.16 & 92.96 & 95.21\\
	\textbf{F$^2$C}   &95.25  &89.38 &99.23 &97.76  &98.25 &92.31 &94.25\\
\bottomrule
\label{tab:6}
\end{tabular}
\end{table*}

\textbf{Ratios within dual data loader.}
We set the ratio of the size between instance-based and identity-based loaders as 1:1 by default.
To further explore the influence of the ratios within dual data loader, we show the experiments in Table \ref{tab:5}.
We utilize ResNet50 as backbone to train MS1MV2 dataset.
We find that the default ratio within dual data loader achieves the highest results on most datasets, especially on challenging IJBC and MegaFace.

\begin{figure}[t]
\centering
\subfloat[Comparison of GPU memory Occupancy (GB).]{\includegraphics[width = .48\linewidth]{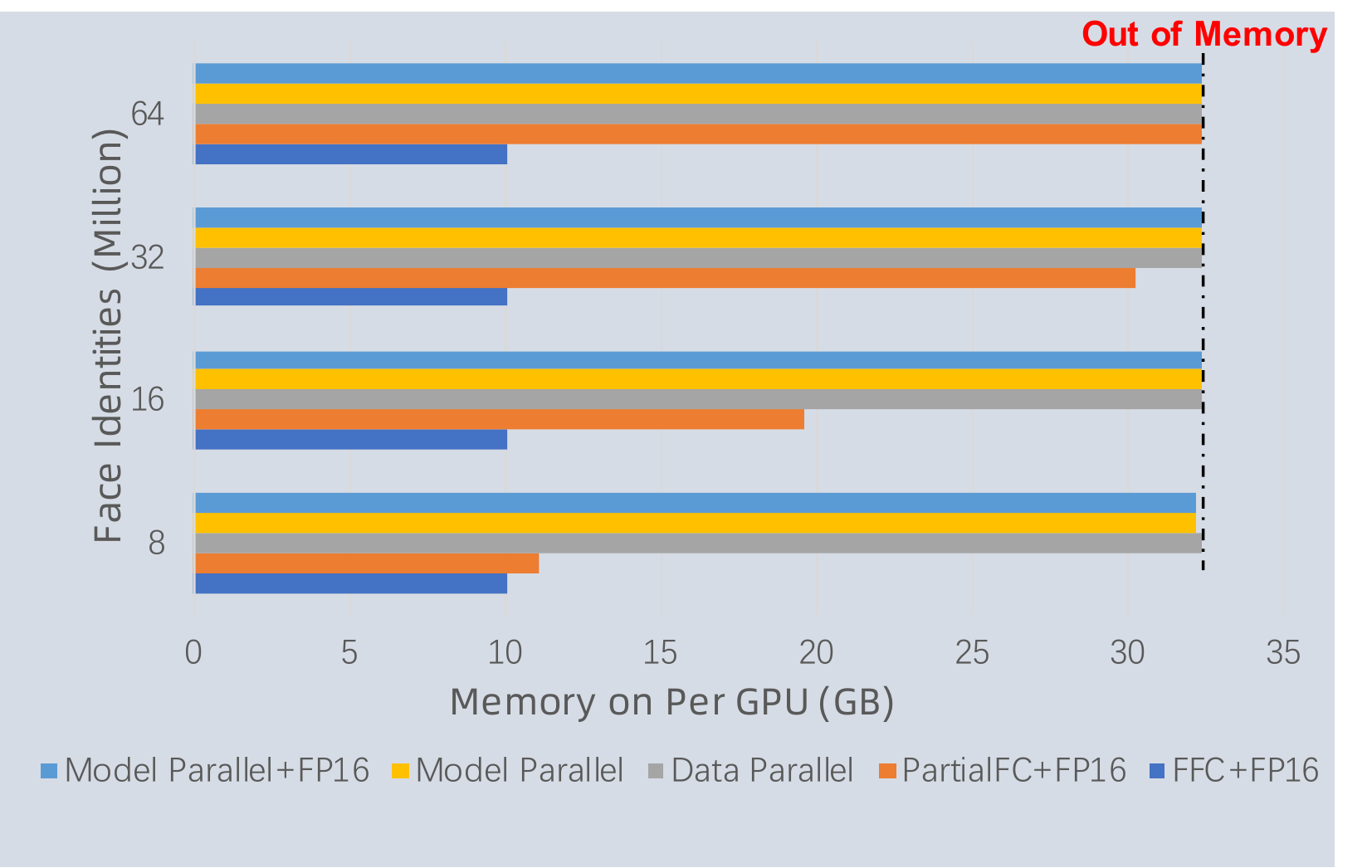}\label{fig:storage}}\hfill
\subfloat[Comparisons of Throughput (Images/Sec.).]{\includegraphics[width = .48\linewidth]{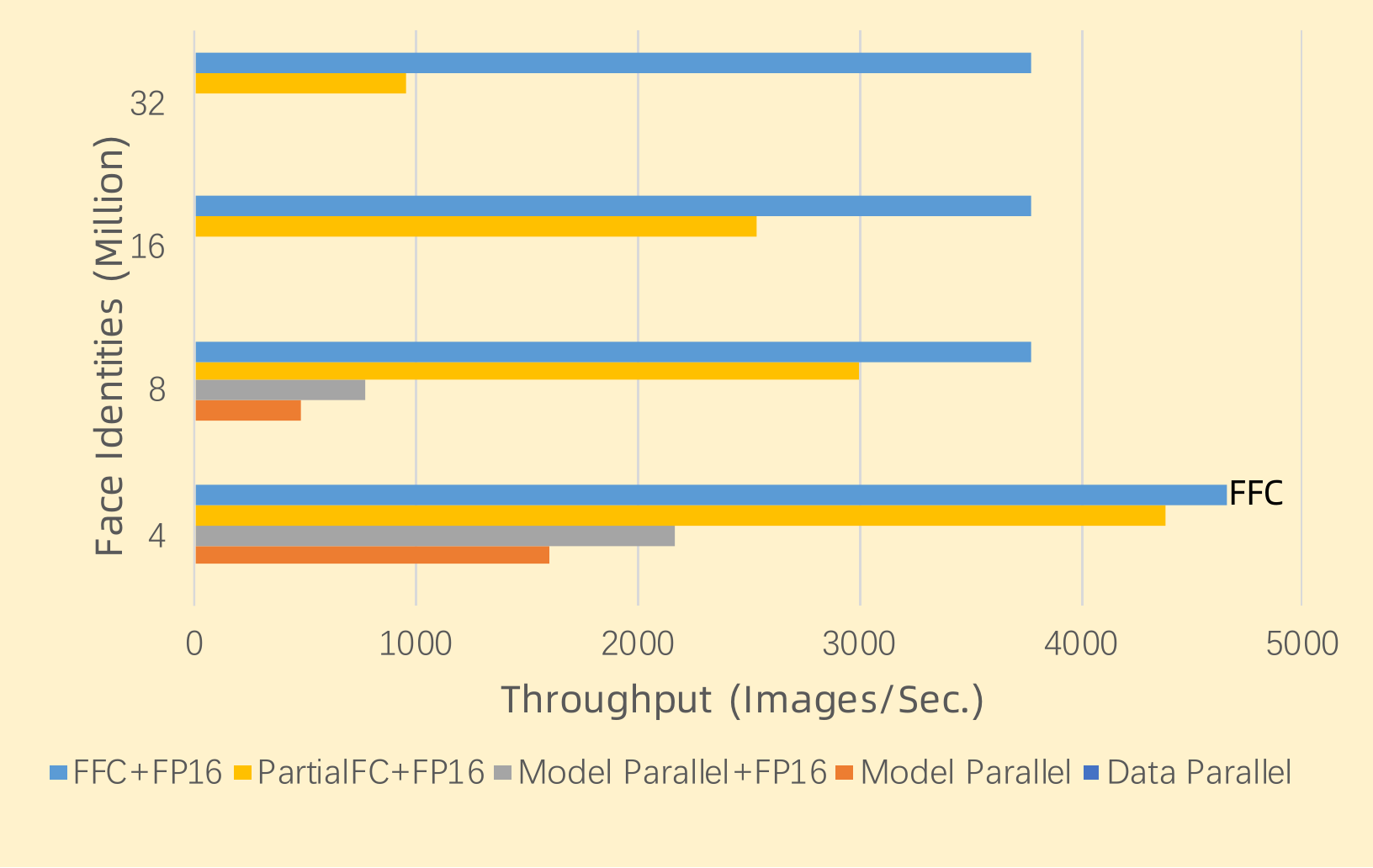}\label{fig:tuntuliang}}
\caption{Visualizations of the hardware resource occupancy of different training methods.}     
\label{fig:hardware_comparison}
\end{figure}

\subsection{Comparisons with SOTA Methods.}
We compare our F$^2$C to other 6 state-of-the-art methods and show the results in Table \ref{tab:6}.
We can observe that F$^2$C outperforms lower-boundary, N-pair, Multi-similarity, and TCP by a large margin, especially on IJBC and MegaFace datasets.
As claimed in VFC \cite{livirtual}, Upper-boundary represents training with normal FC using the 100\% face identities. 
F$^2$C has a little degradation of performance than upper-boundary.
It can also achieve comparable results with Partial-FC, but Partial-FC requires hardware space to store the total identities' centers while VFC doesn't. However the performance of VFC drops obviously compared to F$^2$C.





\textbf{Visualizations of Resource Cost and Training Efficiency.}
The GPU memory occupancy and throughput are two crucial factors to evaluate the practicability of a method in distributed parallel training.
To better understand the efficiency of F$^2$C, Figure \ref{fig:hardware_comparison} visualizes the GPU memory occupancy and throughput of F$^2$C and other training methods. The results are obtained on a 8 V100 32G GPUs.
GPU memory occupancy is illustrated in Figure \ref{fig:storage}, Data Parallel and Model Parallel are out-of-memory (OOM) when the identities reach to 16 millions.
The memory of Partial-FC increases with growth of the identities and it also OOM when the identities reach to 32 millions.
Besides, we show the throughput comparisons in Figure \ref{fig:tuntuliang}, only F$^2$C can keep the high-level throughput among different number of identities.
Therefore, the proposed F$^2$C is practical in ultra-large-scale face recognition task.

\section{Conclusion}
\label{conclusion}
In this paper, we propose an efficient training approach $\text{F}^{2}\text{C}$ for ultra-large-scale face recognition training, the main innovation is Dynamic Class Pool (DCP) for store and update of face identities' feature as an substitute of FC and dual loaders for helping DCP update efficiently. 
The results of comprehensive experiments and analysis show that our approach can reduce hardware cost and time for training as well as obtaining comparable performance to state-of-the-art FC-based methods.

\textbf{Broader impacts.} The proposed method is validated on face training datasets due to the wide variety, the scheme could be expanded to other datasets and situations . However, it does not contain any studies involving affecting ethics or human rights performed by any of the authors.

\section{Acknowledge}
This research is supported by the National Research Foundation, Singapore under its AI Singapore Programme (AISG Award No: AISG2-PhD-2021-08-008). This work is supported by Alibaba Group through Alibaba Research Intern Program. We thank Google TFRC for supporting us to get access to the Cloud TPUs. We thank CSCS (Swiss National Supercomputing Centre) for supporting us to get access to the Piz Daint supercomputer. We thank TACC (Texas Advanced Computing Center) for supporting us to get access to the Longhorn supercomputer and the Frontera supercomputer. We thank LuxProvide (Luxembourg national supercomputer HPC organization) for supporting us to get access to the MeluXina supercomputer. This work is also supported by the Chinese National Natural Science Foundation Projects 62106264 and National Natural Science Foundation of China (62176165).

{\small
\bibliographystyle{ieee_fullname}
\bibliography{references}
}

\end{document}